\documentclass[11pt, a4paper, logo, copyright, nonumbering]{astribot}
\usepackage[authoryear, sort&compress, round]{natbib}
\usepackage{dblfloatfix}
\usepackage{ulem}
\usepackage{caption}
\usepackage{dramatist}
\usepackage{xspace}
\usepackage{pifont} 
\usepackage{multirow}
\usepackage{tcolorbox}
\usepackage{xltabular}
\usepackage{longtable}
\usepackage{hyperref}
\usepackage{array}      
\usepackage{amsfonts}
\usepackage{amsmath}
\usepackage{amssymb}
\usepackage{lineno}
\usepackage{multirow}
\usepackage{adjustbox}
\usepackage{siunitx}
\usepackage{mathtools}
\usepackage{multirow}
\usepackage[bottom]{footmisc}
\usepackage{fontawesome5}
\usepackage{subcaption}
\usepackage{setspace}
\usepackage{lipsum} 
\usepackage{multicol}
\usepackage{annotate-equations}
\usepackage{pdfpages}
\usepackage{threeparttable}
\usepackage{makecell}
\usepackage{xcolor}
\usepackage{epigraph}
\usepackage{longtable}

\makeatletter
\def\@BTrule[#1]{%
  \ifx\longtable\undefined
    \let\@BTswitch\@BTnormal
  \else\ifx\hline\LT@hline
    \nobreak
    \let\@BTswitch\@BLTrule
  \else
     \let\@BTswitch\@BTnormal
  \fi\fi
  \global\@thisrulewidth=#1\relax
  \ifnum\@thisruleclass=\tw@\vskip\@aboverulesep\else
  \ifnum\@lastruleclass=\z@\vskip\@aboverulesep\else
  \ifnum\@lastruleclass=\@ne\vskip\doublerulesep\fi\fi\fi
  \@BTswitch}
\makeatother

\addto\extrasenglish{
}

 {\begin{list}{}%
         {\setlength{\leftmargin}{#1}}%
         \item[]%
 }
 {\end{list}}
 
\renewcommand{\today}{}

\title{Be My Eye : Learning to See \\Beyond a Robot's Own Perspective
}

\author[*]{
Astribot Team
\\
\small
\texttt{research@astribot.com}
\\
\vspace{2em}
\small
Project Page: \href{https://www.astribot.com/research/}{www.astribot.com/AI/CoMind}
\\
\vspace{1em}
\small
Author List in \hyperref[sec:contribution]{Contributions}
}

\begin{abstract}
Robots operating in the real world are inherently subject to partial observability. This limitation is particularly pronounced in multi-robot collaboration, where an individual robot's observations may be insufficient for effective task execution. A central challenge is therefore how to integrate complementary observations from the environment, particularly partner robots, to inform collaborative decision-making. We argue that inter-robot information sharing should go beyond passively transmitting raw observations or generic visual features. Instead, robots should selectively extract and encode task-relevant information for sharing based on their task objectives and current execution progress. To this end, we propose a \textbf{task-oriented active context sharing } framework that encodes robot observations into compact context representations, preserving information critical to collaborative decision-making under a constrained communication budget. Our framework adopts a two-stage training paradigm. In the first stage, onboard robot observations and task instructions are jointly encoded into compact context representations. Task-oriented future prediction and soft task-progress correspondence learning guide this collaboration-agnostic compression process to preserve information relevant to individual robot's task objectives, state transitions, and execution progress. In the second stage, the learned context representations are integrated into a pretrained Vision-Language-Action (VLA) policy, enabling bidirectional context exchange between robots. Each robot integrates complementary information from its partner with its local observations to guide action generation. Supervision of the resulting actions, in turn, jointly refines the shared representations and the policy, aligning information sharing with the requirements of collaborative decision-making.
Our controlled experiments demonstrate substantial performance gains under constrained communication budgets, while three real-world experiments further validate the approach in complex dual-robot collaborative tasks.
These results highlight a shift from passive visual information transmission toward active, task-oriented context sharing, in which compact representations are learned to directly support collaborative decision-making.
\end{abstract}

\begin{document}
\maketitle
\section{Introduction}
\label{sec:intro}

\epigraph{``Gaze following into joint attention, social manipulation into cooperative communication, group activity into collaboration.''}{--- \textup{Michael Tomasello and Malinda Carpenter}, \textit{Shared Intentionality}}

Partial observability is a fundamental constraint on perception in the physical world. For robots, limited fields of view, sensor placement, spatial constraints, and environmental occlusions restrict access to the information needed to execute complex tasks. In multi-robot settings, this limitation also creates an opportunity: robots can exploit complementary observations of their shared environment. Consider two robots collaboratively transporting a large cabinet through a narrow passage. The cabinet may obscure critical regions, leaving obstacles or clearance visible to one robot but hidden from the other. These observations can directly affect whether its partner should slow down, adjust its posture, or change direction. Similar dependencies arise in warehousing, assembly, and everyday collaborative manipulation, where information distributed across robots jointly informs task execution. We argue that the potential of multi-robot systems lies not only in their collective physical capabilities but also in the complementary perspectives afforded by distributed sensing. Translating this perceptual complementarity into coordinated action requires each robot to integrate the observations of its partners into its own decision-making, a fundamental challenge and opportunity in tightly coupled multi-robot collaboration.

A common approach to multi-robot coordination relies on centralized planning, in which a high-level planner decomposes global objectives into subtasks or commands that individual robots execute using local observations. Although effective for task allocation and high-level coordination, systems that communicate primarily through task descriptions or discrete instructions provide limited support for sharing fine-grained perceptual information. Object states, spatial relationships, and execution progress observed by one robot may therefore remain unavailable to another robot's action policy. This limitation is particularly pronounced in dynamic, partially observable environments, where effective collaboration requires continuous adaptation to environmental changes and partners' actions. Without access to complementary perceptual information, coordination is often restricted to parallel task execution or sequential handoffs, rather than tightly coupled behaviors grounded in shared situational awareness.

Conditioning a robot's policy on its partners' visual observations provides a natural mechanism for coupling perception and action across robots. However, transmitting raw images or generic dense visual features incurs substantial communication and computational costs and introduces redundant or task-irrelevant information. Moreover, the value of an observation depends on task objectives, robot roles, and execution progress: visual details essential to one decision may be irrelevant to another. Generic visual compression may preserve appearance or semantic content without retaining the information most useful for collaborative action. The central challenge is therefore to selectively encode local observations into compact representations that preserve their relevance to collaborative decision-making. We refer to this process as \textbf{active context sharing}. Here, \textit{active} denotes task-conditioned information compression, rather than selection of communication timing or recipients; these decisions may still be managed by a centralized planner. Active context sharing aims to preserve task-relevant perceptual content together with relationships among task objectives, observed states, and execution progress, providing partners with informative context under a limited representation budget.

In this work, we propose a \textbf{two-stage active context sharing framework} for multi-robot collaboration.
In Stage~1, we introduce a \textbf{task-oriented context compression} module that uses learnable queries to aggregate frozen multi-camera visual features into compact latent tokens conditioned on task instructions. We train this module with two complementary objectives to preserve task-relevant visual information and capture the structure of task execution:
(i) task-oriented future prediction encourages the latent tokens to encode information predictive of future visual states, emphasizing task-relevant regions and state transitions; and
(ii) soft task-progress correspondence learning structures the latent space according to task identity and execution progress, promoting inter-task discriminability and intra-task temporal continuity.
In Stage~2, we perform \textbf{joint context optimization and collaborative policy learning}. Context tokens produced by the Stage~1 encoder are projected into the hidden space of a pretrained Vision-Language-Action (VLA) policy and integrated into its multi-modal prefix, conditioning action generation alongside local visual observations, task instructions, and proprioceptive states. This interface supports variable numbers of shared tokens without modifying the action-generation architecture. Joint optimization of the context encoder and collaborative policy allows the receiving robot's action supervision to refine the sending robot's representations for collaborative decision-making. Symmetric encoding and policy interfaces enable bidirectional sharing, allowing each robot to incorporate complementary context from its partners.

We systematically evaluate our framework across diverse real-world dual-robot manipulation tasks involving visual occlusion, spatial separation, state-dependent interactions, and role coordination, all of which require robots to leverage complementary observations for successful execution. We compare our approach against communication-free policies and alternative information-sharing strategies under varying communication budgets, while conducting ablation studies to assess the contribution of key design choices.

Our main contributions are as follows:
\begin{enumerate}
    \item \textbf{Active context sharing framework for multi-robot collaboration.}
    We propose a two-stage training framework that encodes each robot's local observations into compact, task-conditioned latent representations, which are integrated into partner robots' action policies and jointly optimized through downstream action supervision. This approach complements high-level coordination with shared perceptual context, enabling information from complementary observations to directly inform collaborative action generation.
    \item \textbf{Progress-aware, task-conditioned observation compression.}
    We introduce a query-based compression module supporting joint and per-camera visual encoding. Task-oriented future prediction and soft task-progress correspondence learning jointly encourage the retention of predictive visual information, inter-task discriminability, and intra-task temporal continuity under a fixed representation budget.
    \item \textbf{Joint context optimization for collaborative policy learning.}
    We integrate compressed context into pretrained VLA policies through a flexible conditioning interface supporting variable token budgets and bidirectional context sharing. Action supervision from the receiving robot jointly optimizes the context encoder and policy, aligning shared representations with the requirements of collaborative decision-making.
    \item \textbf{Real-world evaluation and representation analysis.}
    We systematically evaluate our framework on dual-robot manipulation tasks through comparisons against no-sharing and alternative information-sharing baselines, communication-budgets, and component-wise ablation studies.
\end{enumerate}

\section{Method}

\begin{figure*}[t]
  \centering
  \includegraphics[width=\textwidth]{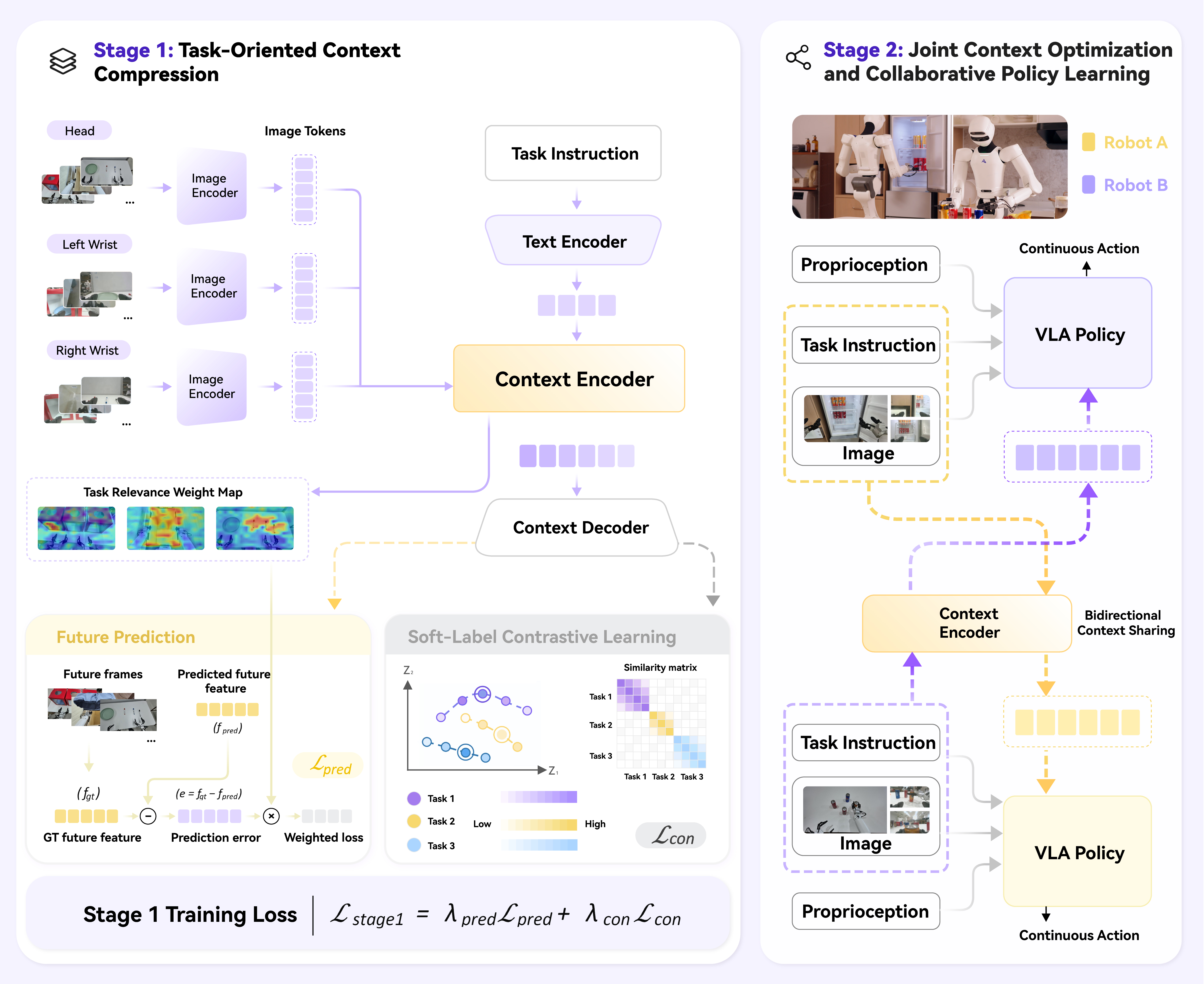}
  \caption{\textbf{Overview of the two-stage active context sharing framework.} In Stage 1, multi-camera observations and task instructions are compressed into shared context, optimized through weighted future visual feature prediction and soft-label contrastive learning. In Stage 2, robots bidirectionally exchange local context and condition their VLA policies on shared context to generate continuous actions for collaborative decision-making.}
  \label{fig:framework}
\end{figure*}

\subsection{Problem Formulation and Framework Overview}
We study \textbf{active context sharing} for multi-robot collaboration, where robots exchange task-relevant latent representations to complement each other's local observations and improve collaborative decision-making. Throughout this paper, $i,j$ denote robot indices, $t$ denotes the time step, $c$ denotes the camera index, $p$ denotes the visual patch index, and $n,m$ denote sample indices within a training batch.

At time $t$, robot $i$ receives local visual observations from $C$ cameras:
\begin{equation}
\mathbf{V}_{i,t}
=
\left\{
\mathbf{V}_{i,t}^{(c)}
\right\}_{c=1}^{C},
\qquad
\mathbf{V}_{i,t}^{(c)}
\in
\mathbb{R}^{H\times W\times 3},
\label{eq:multicam-observation}
\end{equation}
where $H$ and $W$ denote the image height and width, respectively, along with a language instruction $\ell_i$ and proprioceptive state $\mathbf{s}_{i,t}$.

Due to limited fields of view, spatial constraints, and environmental occlusions, local observations may not provide sufficient information for collaborative task execution. To address this, a collaborating robot $j$ encodes its multi-camera observations and task instruction into a task-conditioned latent representation $\mathbf{Z}_{j,t}$, which serves as additional context for robot $i$'s Vision-Language-Action (VLA) policy:
\begin{equation}
\mathbf{A}_{i,t}
=
\pi_{\theta}
\left(
\mathbf{V}_{i,t},
\ell_i,
\mathbf{s}_{i,t},
\mathbf{Z}_{j,t}^{proj}
\right),
\qquad
\mathbf{A}_{i,t}
\in
\mathbb{R}^{H_a\times d_a},
\label{eq:action-policy}
\end{equation}
where $\pi_{\theta}$ denotes the VLA policy, $\mathbf{Z}_{j,t}^{proj}$ represents the shared latent context projected into the policy's hidden space, and $H_a$ and $d_a$ denote the action prediction horizon and per-step action dimension, respectively.

We refer to the task-conditioned encoding of local observations into shared latent representations as \textbf{active context sharing}. Here, \textit{active} characterizes how shared information is constructed, rather than autonomous decisions about when or with whom to communicate.

Our framework adopts symmetric, bidirectional latent sharing. For two collaborating robots, A and B, their action predictions are given by
\begin{equation}
\mathbf{A}_{A,t}
=
\pi_{\theta}
\left(
\mathbf{V}_{A,t},
\ell_A,
\mathbf{s}_{A,t},
\mathbf{Z}_{B,t}^{proj}
\right),
\label{eq:action-policy-a}
\end{equation}
\begin{equation}
\mathbf{A}_{B,t}
=
\pi_{\theta}
\left(
\mathbf{V}_{B,t},
\ell_B,
\mathbf{s}_{B,t},
\mathbf{Z}_{A,t}^{proj}
\right).
\label{eq:action-policy-b}
\end{equation}
Each robot thus both contributes to and benefits from shared contextual information, without requiring predefined leader--follower roles.

Our framework adopts a \textbf{two-stage training paradigm}. In Stage~1, task-conditioned context tokens are optimized to preserve information relevant to task objectives and state transitions under a limited representation budget, while promoting inter-task discriminability and intra-task temporal continuity. In Stage~2, these tokens are integrated into a pretrained VLA policy, and downstream action supervision jointly refines the shared context representations and policy for collaborative decision-making.

\subsection{Task-Oriented Multi-Camera Context Encoding}

Given a set of multi-camera observations
$\mathbf{V}_{i,t} = \{\mathbf{V}_{i,t}^{(c)}\}_{c=1}^{C}$
and a task instruction $\ell_i$, our objective is to encode the local
observations of robot $i$ into a compact context representation that
preserves information relevant to the current task. Frozen visual and language encoders extract generic feature representations from the observations and task instructions, respectively:
\begin{equation}
\mathbf{F}_{i,t}^{(c)}
= f_{\mathrm{vis}}(\mathbf{V}_{i,t}^{(c)}),
\qquad
\mathbf{L}_i = f_{\mathrm{text}}(\ell_i),
\label{eq:input-encoding}
\end{equation}
where $\mathbf{F}_{i,t}^{(c)} \in \mathbb{R}^{N_p \times d_f}$
and $\mathbf{L}_i \in \mathbb{R}^{N_\ell \times d_\ell}$
denote the visual and linguistic features, respectively.

To extract task-relevant information under a fixed representation budget, a trainable context encoder $E_\phi$ aggregates visual and linguistic features using $K$ learnable queries $\mathbf{Q} \in \mathbb{R}^{K \times d_z}$. The resulting task-conditioned context tokens are given by:
\begin{equation}
\mathbf{Z}_{i,t}
=
E_\phi
\left(
\mathbf{Q},
\{\mathbf{F}_{i,t}^{(c)}\}_{c=1}^{C},
\mathbf{L}_i
\right),
\qquad
\mathbf{Z}_{i,t} \in \mathbb{R}^{K \times d_z}.
\label{eq:task-context-encoding}
\end{equation}

The context encoder supports both joint and per-camera query allocation. Joint encoding uses a shared set of queries to aggregate observations from all cameras into $K$ context tokens. Per-camera encoding assigns $K_c$ queries to each camera $c$ and concatenates the resulting tokens, yielding a total token budget of:
\begin{equation}
K = \sum_{c=1}^{C} K_c.
\label{eq:total-query-count}
\end{equation}
In both configurations, robots exchange the resulting context representation $\mathbf{Z}_{i,t}$ rather than raw visual observations or generic visual features.

Finally, a learnable projection module $P_\rho$ maps the context tokens into the hidden representation space of the downstream VLA policy:
\begin{equation}
\mathbf{Z}_{i,t}^{proj}
=
P_\rho(\mathbf{Z}_{i,t}),
\qquad
\mathbf{Z}_{i,t}^{proj}
\in \mathbb{R}^{K \times d_\pi},
\label{eq:latent-projector}
\end{equation}
where $d_\pi$ denotes the hidden dimension of the VLA policy.
The projected context tokens serve as additional conditioning during policy learning, supporting collaborative decision-making.

\subsection{Stage 1: Collaboration-Agnostic Context Compression}
Stage 1 focuses on optimizing a compact, task-oriented representation of each robot's local observations under a fixed token budget, without assuming or modeling any specific collaboration pattern. We achieve this through two complementary objectives: (i) task-oriented future feature prediction, which encourages the representation to preserve predictive information about task-relevant visual state transitions, and (ii) soft-label contrastive learning, which structures the representation space according to task identity and execution-stage proximity. Together, these objectives yield compact, generalizable representations that capture task-relevant dynamics and preserve meaningful relationships across tasks and execution stages, providing a reusable basis for downstream collaborative policy learning.

\subsubsection{Task-Oriented Future Prediction}

Given a temporal offset $\Delta$, we define the future visual feature target for camera $c$ as
\begin{equation}
\mathbf{F}_{i,t+\Delta}^{(c)}
=
f_{\mathrm{vis}}
\left(
\mathbf{V}_{i,t+\Delta}^{(c)}
\right),
\qquad
\mathbf{F}_{i,t+\Delta}^{(c)}
\in
\mathbb{R}^{N_p\times d_f}.
\label{eq:future-target}
\end{equation}
The frozen visual encoder maps future observations to a fixed feature space, providing self-supervised prediction targets without exposing future observations to the context encoder.

A predictor $D_{\psi}$ estimates future visual features for all cameras from the current context tokens: 
\begin{equation}
\widehat{\mathbf{F}}_{i,t+\Delta}
=
D_{\psi}(\mathbf{Z}_{i,t})
=
\left\{
\widehat{\mathbf{F}}_{i,t+\Delta}^{(c)}
\right\}_{c=1}^{C}.
\label{eq:future-prediction}
\end{equation}

To prioritize task-relevant visual content, we compute patch-level relevance scores using a pretrained vision-language model with aligned visual and textual representations. Specifically, we measure the similarity between each visual patch and the corresponding task instruction in the aligned embedding space. The resulting similarity scores are normalized to [0, 1] and converted into bounded weights for future visual feature prediction. Let
\begin{equation}
r_n^{(c,p)}\in[0,1]
\label{eq:normalized-relevance}
\end{equation}
denote the normalized relevance of patch $p$ from camera $c$ for training sample $n$. We map these scores to bounded weights:
\begin{equation}
w_n^{(c,p)}
=
w_{\min}
+
(w_{\max}-w_{\min})r_n^{(c,p)},
\label{eq:task-relevance-weight}
\end{equation}
where $0\leq w_{\min}\leq w_{\max}$. The weights are detached from the computational graph during prediction-loss computation, preventing gradients from this objective from propagating through the relevance-scoring branch.

Let $\Omega_n$ denote the set of valid camera-patch pairs for sample $n$. The normalized squared prediction error for each patch is:
\begin{equation}
e_n^{(c,p)}
=
\frac{1}{d_f}
\left\|
\widehat{\mathbf{F}}_n^{(c,p)}
-
\mathbf{F}_n^{(c,p)}
\right\|_2^2.
\label{eq:future-patch-error}
\end{equation}
For a batch of $B_{\mathrm{pred}}$ valid samples, the task-relevance-weighted prediction objective is:
\begin{equation}
\mathcal{L}_{\mathrm{pred}}
=
\frac{1}{B_{\mathrm{pred}}}
\sum_{n=1}^{B_{\mathrm{pred}}}
\frac{
\displaystyle
\sum_{(c,p)\in\Omega_n}
w_n^{(c,p)}e_n^{(c,p)}
}{
\displaystyle
\sum_{(c,p)\in\Omega_n}
w_n^{(c,p)}
}.
\label{eq:future-loss}
\end{equation}
This objective assigns greater predictive importance to task-relevant patches, encouraging the context encoder to preserve information predictive of task-relevant visual state transitions under a fixed token budget.

\subsubsection{Soft Task-Progress Correspondence Learning}
\begin{figure*}[t]
  \centering
  \includegraphics[width=\textwidth]{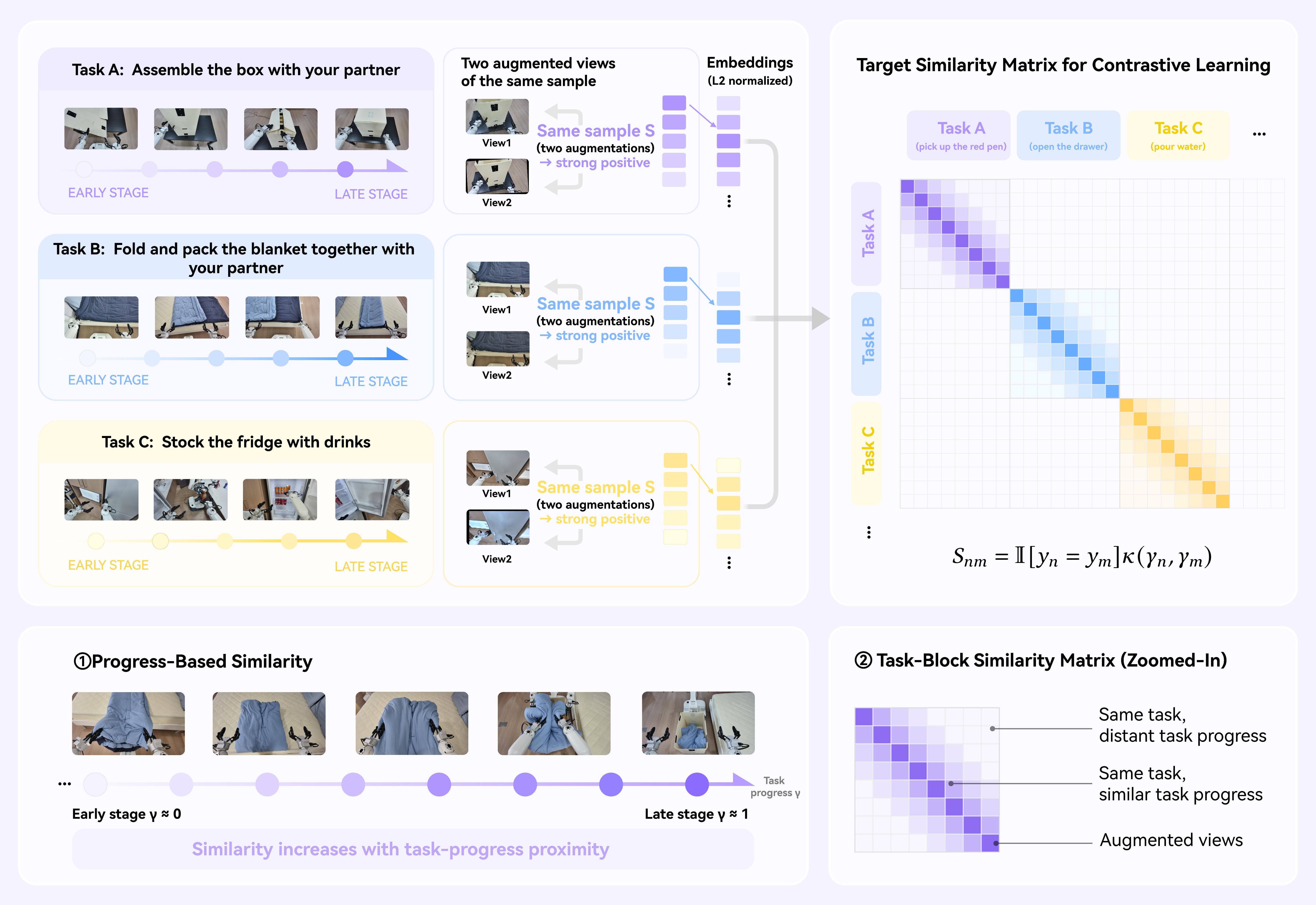}
  \caption{\textbf{Soft Task-Progress Correspondence Learning.} For each observation, augmented views are encoded by the task-conditioned context encoder and treated as positive pairs. Task- and progress-based affinities define the similarity matrix for contrastive learning, promoting cross-view consistency, inter-task discriminability, and intra-task temporal continuity.}
  \label{fig:contrastive}
\end{figure*}

To impose additional structure on the context representation space, we construct soft correspondence targets based on task identity and execution progress. Given a multi-camera observation $\mathbf{V}_n$, we generate two independently augmented views:
\begin{equation}
\widetilde{\mathbf{V}}_n^{(v)}
=
\mathcal{T}_v(\mathbf{V}_n),
\qquad
v\in\{1, 2\},
\label{eq:augmented-views}
\end{equation}
where $\mathcal{T}_v$ denotes a stochastic vision augmentation operator. All augmented views are encoded using the same task instruction and context encoder:
\begin{equation}
\mathbf{Z}_n^{(v)}
=
E_{\phi}
\left(
\mathbf{Q},
\widetilde{\mathbf{F}}_n^{(v)},
\mathbf{L}_n
\right),
\label{eq:contrastive-latents}
\end{equation}
where $\widetilde{\mathbf{F}}_n^{(v)}$ denotes the visual features extracted from the augmented observations $\widetilde{\mathbf{V}}_n^{(v)}$.

Normalized contrastive embeddings are obtained by pooling the context tokens, applying a projection head $g_{\xi}$, and performing $\ell_2$ normalization:
\begin{equation}
\mathbf{h}_n^{(v)}
=
\frac{
g_{\xi}\left(
\operatorname{Pool}(\mathbf{Z}_n^{(v)})
\right)
}{
\left\|
g_{\xi}\left(
\operatorname{Pool}(\mathbf{Z}_n^{(v)})
\right)
\right\|_2
}.
\label{eq:contrastive-representation}
\end{equation}
The projection head is used only for contrastive supervision; the complete context tokens remain the representation exchanged between robots.

Let $B_{\mathrm{con}}$ denote the contrastive batch size. 
For sample $n$, let $y_n$ denote its task identity and $\gamma_n \in [0,1]$ its normalized execution progress. We compute the execution progress as $\gamma_n = t_n/(T_n-1)$, where $t_n$ is the zero-based frame index of sample $n$ and $T_n$ is the total number of frames in its demonstration trajectory.We define the pairwise affinity between samples $n$ and $m$ as:
\begin{equation}
S_{nm}
=
\mathbb{I}[y_n=y_m]
\kappa(\gamma_n,\gamma_m),
\label{eq:soft-relation}
\end{equation}
where $\mathbb{I}[\cdot]$ is the indicator function and $\kappa$ is a nonnegative similarity kernel that decreases with execution-progress distance. Execution progress is used exclusively to construct training targets and is not provided to the context encoder or communicated between robots.

We normalize the affinity matrices row-wise to obtain directional target distributions:
\begin{equation}
\mathbf{R}^{1\rightarrow2}
=
\operatorname{RowNorm}(\mathbf{S}),
\qquad
\mathbf{R}^{2\rightarrow1}
=
\operatorname{RowNorm}(\mathbf{S}^{\top}).
\label{eq:bidirectional-progress-targets}
\end{equation}
Here, $\operatorname{RowNorm}(\cdot)$ denotes row-wise normalization, with a suitable safeguard for rows whose affinity sums are zero.

To balance instance-level correspondence with progress-based similarity, we define the soft targets as:
\begin{equation}
\mathbf{P}^{d}
=
\alpha\mathbf{I}
+
(1-\alpha)\mathbf{R}^{d},
\qquad
d\in\{1\rightarrow2,2\rightarrow1\},
\label{eq:soft-target}
\end{equation}
where $\alpha\in[0,1]$ controls the relative contribution of the two components. Since both $\mathbf{I}$ and $\mathbf{R}^{d}$ are row-stochastic, $\mathbf{P}^{d}$ is also row-stochastic.

The predicted cross-view correspondence distributions are computed using temperature-scaled cosine similarities:
\begin{equation}
Q_{nm}^{1\rightarrow2}
=
\frac{
\exp\left(
(\mathbf{h}_n^{(1)})^\top
\mathbf{h}_m^{(2)}/\tau
\right)
}{
\displaystyle
\sum_{j=1}^{B_{\mathrm{con}}}
\exp\left(
(\mathbf{h}_n^{(1)})^\top
\mathbf{h}_j^{(2)}/\tau
\right)
},
\label{eq:predicted-relation-forward}
\end{equation}
\begin{equation}
Q_{nm}^{2\rightarrow1}
=
\frac{
\exp\left(
(\mathbf{h}_n^{(2)})^\top
\mathbf{h}_m^{(1)}/\tau
\right)
}{
\displaystyle
\sum_{j=1}^{B_{\mathrm{con}}}
\exp\left(
(\mathbf{h}_n^{(2)})^\top
\mathbf{h}_j^{(1)}/\tau
\right)
},
\label{eq:predicted-relation-backward}
\end{equation}
where $\tau>0$ is the contrastive temperature.

The bidirectional soft-label contrastive objective is defined as:
\begin{equation}
\mathcal{L}_{\mathrm{con}}
=
-\frac{1}{2B_{\mathrm{con}}}
\sum_{d\in\{1\rightarrow2,2\rightarrow1\}}
\sum_{n=1}^{B_{\mathrm{con}}}
\sum_{m=1}^{B_{\mathrm{con}}}
P_{nm}^{d}\log Q_{nm}^{d}.
\label{eq:soft-contrastive-loss}
\end{equation}
This objective encourages consistent representations across augmented views while aligning samples according to task identity and execution-stage proximity. Samples from the same task and nearby execution stages receive greater target affinity, whereas samples from different tasks are assigned zero affinity before instance-level mixing. The resulting supervision promotes task-discriminative representations with structured temporal relationships.

The overall Stage 1 objective is:
\begin{equation}
\mathcal{L}_{\mathrm{S1}}
=
\lambda_{\mathrm{pred}}\mathcal{L}_{\mathrm{pred}}
+
\lambda_{\mathrm{con}}\mathcal{L}_{\mathrm{con}},
\label{eq:stage1-objective}
\end{equation}
where $\lambda_{\mathrm{pred}}$ and $\lambda_{\mathrm{con}}$ control the relative contributions of future feature prediction and soft-label contrastive learning, respectively.

\subsection{Stage 2: Joint Context Optimization and Collaborative Policy Learning}

For illustration, we consider robot $j$ as the context sender and robot $i$ as the context receiver. Following Stage 1, the context encoder generates task-oriented context tokens from the sender's current multi-camera observations and task instruction:
\begin{equation}
\mathbf{Z}_{j,t}
=
E_{\phi}
\left(
\mathbf{Q},
\mathbf{F}_{j,t},
\mathbf{L}_{j}
\right).
\label{eq:stage2-sender-latent}
\end{equation}
A projection module then maps the context tokens into the hidden space of the VLA policy:
\begin{equation}
\mathbf{Z}_{j,t}^{proj}
=
P_{\rho}
\left(
\mathbf{Z}_{j,t}
\right).
\label{eq:stage2-projected-latent}
\end{equation}
The receiving robot $i$ integrates the shared tokens with its local multi-camera observations, task instruction, and proprioceptive state as inputs to the VLA policy. Conditioned on these inputs, the policy predicts the action sequence $\mathbf{A}_{i,t}$ as:
\begin{equation}
\mathbf{A}_{i,t}
=
\pi_{\theta}
\left(
\mathbf{V}_{i,t},
\ell_i,
\mathbf{s}_{i,t},
\mathbf{Z}_{j,t}^{proj}
\right).
\label{eq:stage2-policy}
\end{equation}
For each sender--receiver direction $j\rightarrow i$, we define the Stage 2 training objective as:
\begin{equation}
\mathcal{L}_{\mathrm{S2}}^{j\rightarrow i}
=
\mathcal{L}_{\mathrm{VLA}}^{i}
+
\eta
\mathcal{L}_{\mathrm{S1}}^{j},
\label{eq:stage2-objective}
\end{equation}
where $\mathcal{L}_{\mathrm{VLA}}^{i}$ denotes the action learning objective of the receiving robot, $\mathcal{L}_{\mathrm{S1}}^{j}$ denotes the auxiliary Stage 1 objective computed from the sender's observations and task instruction, and $\eta\geq0$ controls the contribution of the auxiliary loss.

During joint optimization, the downstream action loss propagates through the shared context tokens to the sender's encoder, adapting the transmitted representations to the receiver's collaborative action generation. Meanwhile, future visual feature prediction and soft-label contrastive learning serve as auxiliary objectives that preserve task-relevant perceptual information and the structural relationships between task identity and execution progress. Thus, Stage 2 optimizes the representations not merely for perceptual compression, but also for their utility in downstream collaborative decision-making.

We adopt a symmetric parameterization for the two communication directions, $A\rightarrow B$ and $B\rightarrow A$, using shared encoder, projection, and policy parameters. At inference time, the two robots independently generate their task-conditioned context tokens:
\begin{equation}
\mathbf{Z}_{A,t}
=
E_{\phi}
\left(
\mathbf{Q},
\mathbf{F}_{A,t},
\mathbf{L}_{A}
\right),
\qquad
\mathbf{Z}_{B,t}
=
E_{\phi}
\left(
\mathbf{Q},
\mathbf{F}_{B,t},
\mathbf{L}_{B}
\right).
\label{eq:bidirectional-latents}
\end{equation}
Each robot then conditions its policy on the context received from its partner:
\begin{equation}
\mathbf{A}_{A,t}
=
\pi_{\theta}
\left(
\mathbf{V}_{A,t},
\ell_A,
\mathbf{s}_{A,t},
\mathbf{Z}_{B,t}^{proj}
\right),
\label{eq:inference-action-a}
\end{equation}
\begin{equation}
\mathbf{A}_{B,t}
=
\pi_{\theta}
\left(
\mathbf{V}_{B,t},
\ell_B,
\mathbf{s}_{B,t},
\mathbf{Z}_{A,t}^{proj}
\right).
\label{eq:inference-action-b}
\end{equation}
This formulation enables bidirectional context sharing, allowing each robot to supplement its local observations with task-relevant information from its partner's complementary observations. The future feature prediction decoder $D_{\psi}$ and contrastive projection head $g_{\xi}$ are used only during training and are omitted during cross-robot communication and action inference.

\section{Experiments}
\label{sec:exp}

\subsection{Experimental Setup}
\label{sec:exp-setup}
We evaluate the proposed framework on real-world dual-robot manipulation tasks using two Astribot S1 robots~\citep{gao2025towards}. At each decision step, each robot receives its local camera observations, task instruction, and proprioceptive state, which are jointly encoded into task-conditioned context tokens and subsequently exchanged between the two robots. Each robot integrates the received context into a pretrained $\pi_{0.5}$ policy to generate continuous actions. The framework does not impose predefined sender--receiver roles: both robots can simultaneously encode, exchange, and utilize contextual information from their partners, with their respective active and supporting roles emerging dynamically from the current task stage.

During training, demonstrations collected from both robots are consolidated into a unified dataset, enabling the model to learn context encoding and action generation across different role configurations. Unless otherwise specified, all methods share the same data and action pre-processing pipelines, policy initialization, and optimization hyperparameters to ensure a controlled and fair comparison. The default context token budget is set to $K=32$.

\begin{figure*}[t]
  \centering
  \includegraphics[width=\textwidth]{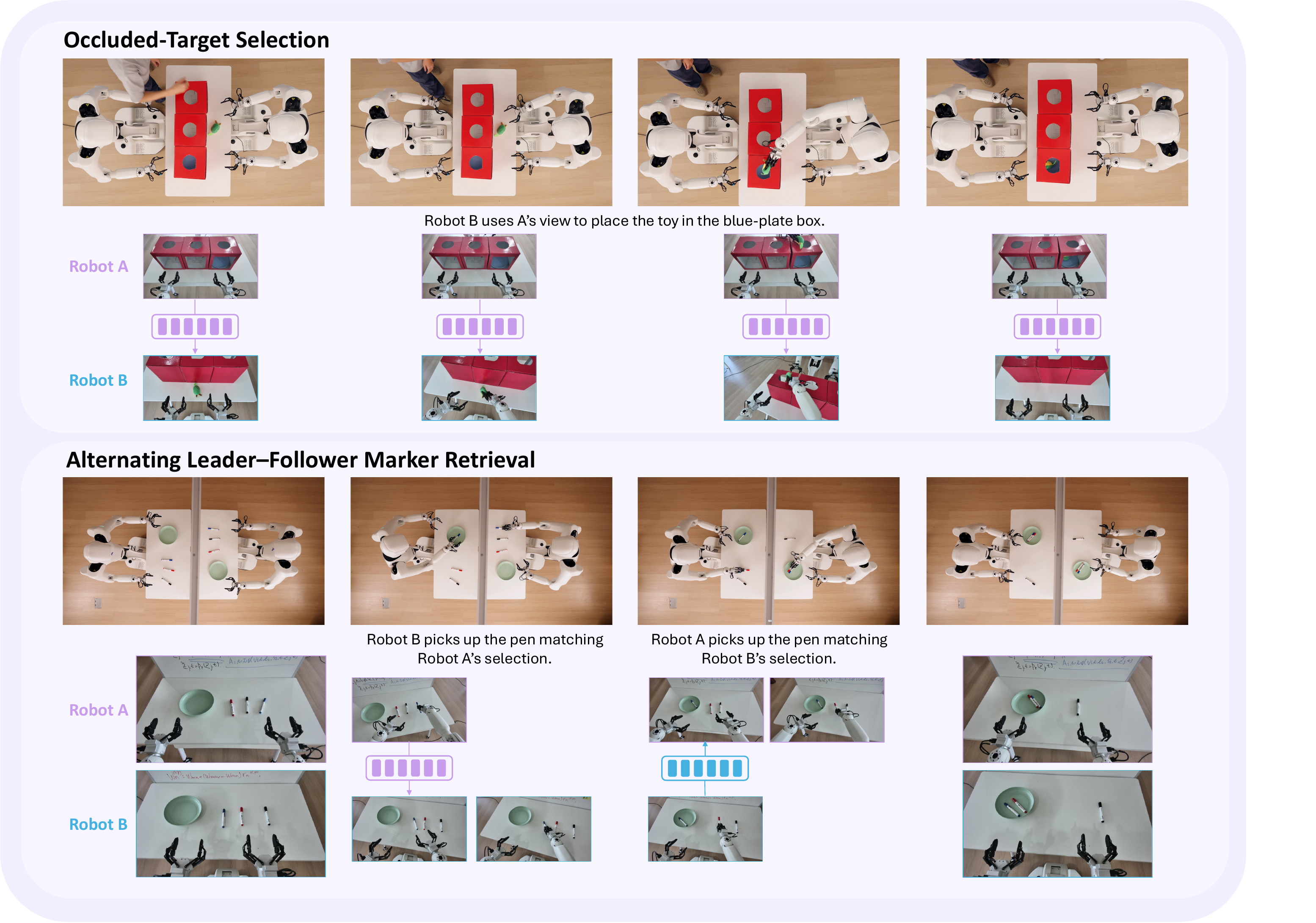}
  \caption{\textbf{Representative rollouts of designed collaborative tasks.} Keyframes highlight critical moments of inter-robot context sharing during task execution.}
  \label{fig:task_illustration}
\end{figure*}

We design two dual-robot tasks to evaluate the effectiveness of cross-robot context sharing. The first is an \textbf{occluded-target selection task}, which evaluates whether shared context tokens can convey target information that is unavailable in the acting robot's local observations. The second is an \textbf{alternating leader-follower marker retrieval task}, which further investigates whether target identity and the partner's execution progress can support bidirectional information sharing and continuous collaborative control. The former minimizes the influence of action coordination, isolating the effectiveness of cross-robot information transfer, whereas the latter requires the two robots to alternate between leader and follower roles within a single task sequence, evaluating whether shared context can directly inform action generation. Representative rollouts are shown in Fig.~\ref{fig:task_illustration}.

The main comparison settings are defined as follows: 
\textbf{(1) No Sharing}, where each robot relies exclusively on its local visual observations, task instruction, and proprioceptive state; 
\textbf{(2) Generic SigLIP}, where the partner's visual tokens are directly shared without task-oriented compression; 
\textbf{(3) Ours-Frozen}, where the context tokens learned in Stage~1 are shared while the context encoder remains frozen during policy training; and 
\textbf{(4) Ours-Joint}, where the context encoder and VLA policy are jointly optimized, allowing downstream action supervision to further shape the information encoded in the shared latent tokens.

\subsection{Occluded-Target Selection}
\label{sec:blind-box}

The occluded-target selection task involves two robots standing face-to-face, with three boxes placed between them. A human repeatedly places a blue plate into one of the boxes. This target-defining action is visible to Robot A but occluded from Robot B. For each experimental setting, we conduct 18 trials.
This task provides a controlled evaluation of whether shared context tokens can convey target information unavailable from the acting robot's local observations. Robot A remains stationary and does not participate in object manipulation, but has access to the visual cues needed to identify the correct target box. Robot B performs the pick-and-place operation, while its local observations alone are insufficient to disambiguate the destination. Successful task completion therefore requires Robot B to leverage the context tokens shared by Robot A to resolve the target ambiguity and select the correct box.

The information-providing robot is intentionally kept stationary to minimize confounding factors arising from action synchronization, physical interaction, and control errors. This task is designed not to evaluate the execution of basic pick-and-place skills, but to isolate whether the shared representations can convey the target-defining information needed by an otherwise capable manipulation policy to select the correct destination among the three available boxes. 

We use target-box selection accuracy as the primary evaluation metric. Table~\ref{tab:blind-box} summarizes the results across different settings. \textbf{Ours-Joint} substantially outperforms \textbf{No Sharing}, demonstrating that compact context tokens effectively convey complementary, target-relevant information beyond the acting robot's local observations. This interpretation is further supported by the behavior under \textbf{No Sharing}: the robot tends to select the nearest box, suggesting that its decisions rely primarily on local spatial cues rather than the target-defining information available from its partner.
To evaluate the effectiveness of our representation against high-bandwidth visual sharing, we further compare \textbf{Ours-Joint} with \textbf{Generic SigLIP}, which serves as a visual-sharing reference with a token budget of 160. Despite using only 32 shared context tokens, \textbf{Ours-Joint} achieves a target selection accuracy of 88.9\%, outperforming \textbf{Generic SigLIP} by 16.7 percentage points. This result suggests that simply transmitting more visual information does not necessarily lead to better collaborative decisions. Instead, task-oriented context compression enables more effective transmission of target-defining information, substantially improving cross-robot information sharing while reducing the token budget by 80\%.

\begin{table}[t]
  \centering
  \caption{Occluded-target selection results.}
  \label{tab:blind-box}
  \begin{tabular}{lc}
    \toprule
    Method
    & Target accuracy (\%)
    \\
    \midrule
    No Sharing          & 38.9\\
    Generic SigLIP      & 72.2 \\
    Ours-Joint     & 88.9\\
    \bottomrule
  \end{tabular}
\end{table}

\subsection{Alternating Leader--Follower Marker Retrieval}
\label{sec:marker-retrieval}

This task is designed to evaluate whether shared context tokens convey both target identity and the partner's execution progress, thereby supporting bidirectional information sharing and continuous collaboration during role switching. Three colored markers (red, blue, and black) are placed in front of each robot, A and B. Each trial consists of two consecutive stages, during which the robots alternate between Leader and Follower roles. Taking Robot A as the initial Leader, in the first stage, Robot A follows its task instruction to select and grasp a marker of the specified color $c_A$. Robot B acts as the Follower and must infer from Robot A's shared context whether the grasp has been completed and when to grasp a marker of the same color. Robot A remains stationary throughout this stage. In the second stage, the roles are reversed. Robot B becomes the Leader and selects and grasps a marker of a different color $c_B$, where $c_B \neq c_A$, from the two remaining colors. Robot A then acts as the Follower, using Robot B's shared context to infer the selected color and determine when to initiate its grasp. A complete sequence initiated by Robot A is denoted as ABBA, representing the order in which the two robots initiate their actions. Importantly, each robot's language instruction specifies only the target color for its own Leader stage. As a Follower, it receives neither an explicit target-color instruction nor an action-trigger command. Instead, it must infer two key pieces of information from its partner's shared context tokens: \textbf{target identity}, i.e., the selected marker color, and \textbf{execution progress}, i.e., whether the partner has completed the grasp and when to initiate its own action. We evaluate all six ordered combinations of distinct marker colors, with both Robot A and Robot B serving as the initial Leader. Each configuration is repeated four times, resulting in 48 trials.

Under \textbf{No Sharing}, the initial Leader successfully grasps the marker specified by its instruction and remains stationary. The initial Follower typically waits for a period before initiating its action. However, without access to its partner's target selection or execution status, it tends to retrieve the color specified in its own subsequent Leader-stage instruction rather than match its partner's current target. This behavior suggests that the policy retains basic grasping capabilities and can learn coarse temporal patterns, such as delayed action initiation, but cannot reliably condition its behavior on the partner's target identity and real-time execution progress.

We evaluate collaboration using four metrics: color-matching accuracy when Robot~B follows Robot~A, color-matching accuracy when Robot~A follows Robot~B, correct-trigger rate, and overall two-stage sequence success rate. A trial is considered successful only when both stages are completed in the correct order and with the corresponding target colors. Together, these metrics assess bidirectional target inference, progress-aware action triggering, and their combined effectiveness in completing the full collaborative sequence.

\paragraph{Normalized Process Score.}
To quantify partial completion of the alternating leader–follower sequence, we assign 0.25 points for each successful Leader grasp, 0.5 points for each correctly triggered Follower action, and 1.0 point for each correctly matched Follower grasp.

\begin{table}[t]
  \centering
  \caption{Alternating leader--follower marker-retrieval results.}
  \label{tab:marker-retrieval}
  \begin{tabular}{lcccc}
    \toprule
    Method
    & B follows A (\%)
    & A follows B (\%)
    & Trigger (\%)
    & Sequence Success Rate (\%)\\
    \midrule
    No Sharing          & 16.7& 33.3& 25& 0\\
    Generic SigLIP      & 100 & 100 & 100 & 100 \\
    Ours-Joint       & 100& 100& 100& 100\\
    \bottomrule
  \end{tabular}
\end{table}

As shown in Table~\ref{tab:marker-retrieval}, \textbf{No Sharing} achieves low color-matching accuracy in both following directions (16.7\% for B follows A and 33.3\% for A follows B), together with a trigger accuracy of only 25\% and zero full-sequence successes. These results reveal that recognizing the task structure and learning coarse temporal patterns are insufficient for reliable role-switching collaboration: the Follower must access both the partner's target identity and execution progress to coordinate its actions correctly. In contrast, both \textbf{Generic SigLIP} and \textbf{Ours-Joint} achieve 100\% across all four metrics, demonstrating reliable bidirectional target inference and progress-aware action triggering. Notably, \textbf{Ours-Joint} matches the performance of \textbf{Generic SigLIP} while using only 32 shared context tokens, compared with 160 visual tokens. This result demonstrates that task-oriented context compression preserves the information necessary for successful bidirectional collaboration with an 80\% reduction in communication budget.

\subsection{Ablation of Stage 1 Training Objectives}
\label{sec:stage1-ablation}

This section investigates the contributions of the two representation-learning objectives, task-oriented future prediction and task-progress contrastive learning, on the Alternating Leader–Follower Marker Retrieval task. We consider two ablation variants: \textbf{w/o Prediction}, which removes future prediction while retaining the complete contrastive objective, and \textbf{w/o Contrastive}, which removes both self-pair correspondence and task-conditioned progress correspondence while retaining future prediction. The full method combines both objectives.

Each variant undergoes Stage~1 pretraining with its corresponding objectives, followed by joint Stage~2 training. The context encoder remains trainable in all settings. During Stage~2, receiver-side action supervision is combined with the auxiliary objectives retained by each variant, while the removed objectives remain disabled. This comparison isolates the contribution of each representation-learning objective within the complete two-stage training pipeline.

\begin{table}[!htbp]
  \centering
  \caption{Ablation of Stage 1 Training Objectives.}
  \label{tab:stage1-ablation}
  \begin{tabular}{lc}
    \toprule
    Setting
    & Process Score (\%)\\
    \midrule
    Full                & 100 \\
    w/o Prediction       & 96.7\\
    w/o Contrastive       & 68.9\\
    \bottomrule
  \end{tabular}
\end{table}

As shown in Table~\ref{tab:stage1-ablation}, removing contrastive learning substantially reduces the Process Score from 100.0\% to 68.9\%, whereas removing future prediction leads to a smaller decrease to 96.7\%. This asymmetry suggests that task-progress contrastive learning plays a more critical role in this task, where successful collaboration requires the Follower to distinguish both the partner's target identity and execution phase. By enforcing task-conditioned correspondence across observations, contrastive learning helps organize shared representations according to task identity and progress. In comparison, future prediction encourages the preservation of predictive information about relevant visual dynamics. Although its individual contribution is smaller, the full method still outperforms the contrastive-only variant, indicating that predictive supervision provides complementary benefits beyond task-progress alignment. Together, the two objectives guide the representation to learn both \textbf{what information to preserve} and \textbf{how to organize it} for collaborative decision-making.

\subsection{Joint Optimization and Communication Capacity}
\label{sec:joint-capacity}

\subsubsection{Frozen Context vs. Joint Optimization}

We further compare \textbf{Ours-Frozen} and \textbf{Ours-Joint} on the Alternating Leader-Follower Marker Retrieval task to investigate whether downstream action supervision can further refine the representations learned in Stage~1. \textbf{Ours-Frozen} keeps the Stage~1 context encoder fixed and trains only the VLA policy conditioned on the shared context tokens. In contrast, \textbf{Ours-Joint} allows action gradients to propagate through the shared context tokens into the context encoder, while retaining the Stage~1 future prediction and progress correspondence objectives. This comparison examines both the transferability of Stage~1 representations to downstream collaborative tasks and the extent to which action supervision can further adapt these representations to support continuous collaborative decision-making.

\begin{table}[!htbp]
  \centering
  \caption{Frozen context vs. joint optimization.}
  \label{tab:frozen-joint}
  \begin{tabular}{lc}
    \toprule
    Setting
    & Process Score (\%)\\
    \midrule
    Ours-Frozen    & 83.3\\
    Ours-Joint  & 100\\
    \bottomrule
  \end{tabular}
\end{table}

As shown in Table~\ref{tab:frozen-joint}, \textbf{Ours-Frozen} achieves a Process Score of 83.3\%, indicating that the representations learned in Stage~1 already capture useful information for downstream collaborative decision-making without further adaptation. Joint Stage~2 training further improves the score to 100\% with a performance gain of 16.7 percentage points. This improvement demonstrates that receiver-side action supervision further adapts the shared context representations to the requirements of collaborative policy execution.

\subsubsection{Context Token Budget}

To investigate the relationship between communication capacity and task performance, we vary the number of shared context tokens, $K \in \{4, 16, 32\}$, on the Alternating Leader-Follower Marker Retrieval task, using \textbf{Generic SigLIP} as a high-bandwidth visual-sharing reference. All token-budget variants of our method share the same feature dimension and training protocol, differing only in the number of transmitted tokens. We evaluate each setting using the Process Score and report the corresponding token count and communication cost.Communication cost is calculated based on the transmitted
token representations. Each token contains 1152
floating-point values stored in FP32 format (4 bytes).
The communication cost per direction and decision step
is therefore computed as $K \times 1152 \times 4$ bytes,
excluding protocol overhead.

\begin{table}[!htbp]
  \centering
  \caption{Context token budget vs. task success rate.}
  \label{tab:token-budget}
  \begin{tabular}{lccc}
    \toprule
    Setting
    & \# Tokens
    & Process Score (\%)& Bytes/step (KiB)\\
    \midrule
    Ours-$K{=}4$& 4& 90.5& 18\\
    Ours-$K{=}16$& 16& 97.6& 72\\
    Ours-$K{=}32$& 32& 100& 144\\
    Generic SigLIP    & 160 & 100& 720\\
    \bottomrule
  \end{tabular}
\end{table}

As shown in Table~\ref{tab:token-budget}, increasing the context token budget consistently improves task performance, revealing a trade-off between communication capacity and collaborative decision-making. Notably, even with only 4 tokens, our method achieves a Process Score of 90.5\%, indicating that task-relevant information can be encoded in a highly compact representation. Increasing the budget to 16 tokens further improves the score to 97.6\%, while 32 tokens yields 100\% performance, matching \textbf{Generic SigLIP} with 80\% fewer transmitted tokens. This progression suggests that additional tokens progressively improve the preservation of decision-relevant information, with diminishing performance gains at larger budgets. We therefore adopt $K=32$ as the default, as it is the smallest evaluated budget that matches the high-bandwidth reference. Overall, these results demonstrate that task-oriented context compression achieves comparable collaborative performance with substantially lower communication overhead.

\subsection{Qualitative Analysis}
\label{sec:qualitative}

\begin{figure*}[!htbp]
  \centering
  \includegraphics[width=\textwidth]{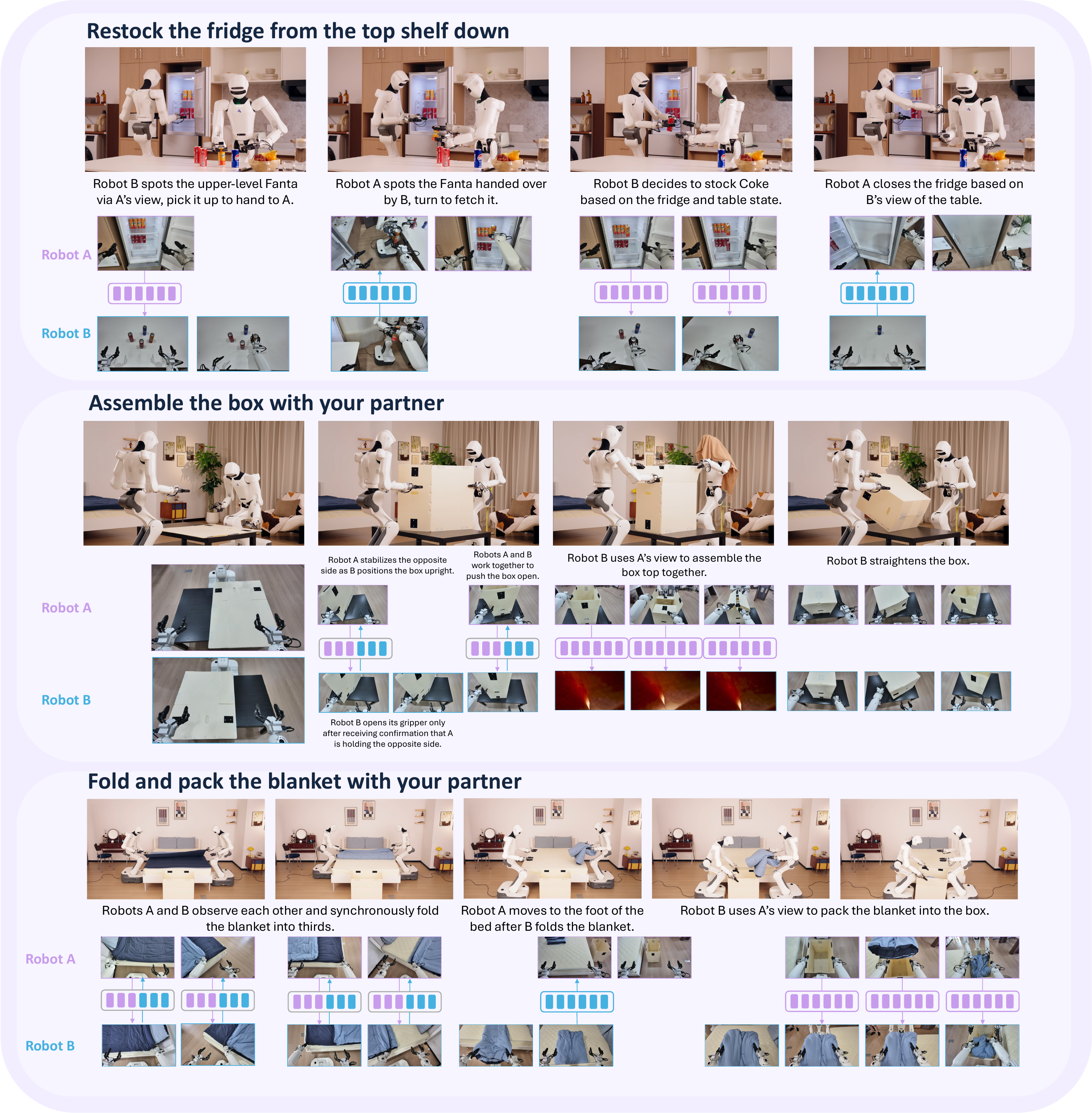}
  \caption{\textbf{Representative rollouts of more collaborative tasks.} Keyframes highlight critical moments of inter-robot context sharing during task execution.}
  \label{fig:more_task_illustration}
\end{figure*}

To further demonstrate the applicability of our approach across different collaborative scenarios, we present three representative dual-robot tasks in Fig.~\ref{fig:more_task_illustration}: (1) refrigerator restocking, illustrating efficient information exchange between robots; (2) box assembly, demonstrating complementary perception and collaborative manipulation under occlusion; and (3) blanket folding and packing, illustrating synchronized dual-robot actions during continuous task execution. These tasks represent different collaborative requirements, including information exchange, occlusion-aware collaboration, and synchronized actions. Representative keyframes illustrate the execution process of each task.

\begin{figure*}[!htbp]
    \centering
    \setlength{\tabcolsep}{2pt}

    \begin{tabular}{c c c c c}
        & \textbf{$t_1$} & \textbf{$t_2$} & \textbf{$t_3$} & \textbf{$t_4$} \\[2pt]

        \rotatebox{90}{\textbf{Robot A}}
        &
        \includegraphics[width=0.215\textwidth]{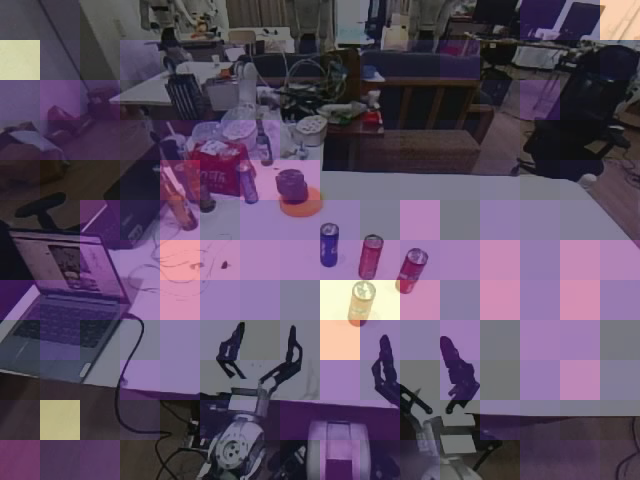}
        &
        \includegraphics[width=0.215\textwidth]{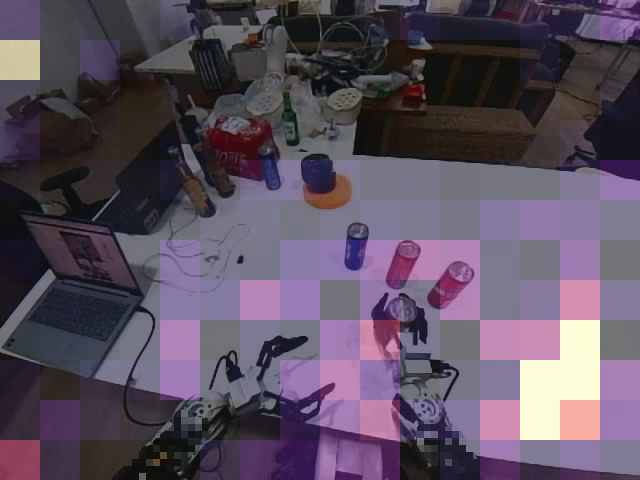}
        &
        \includegraphics[width=0.215\textwidth]{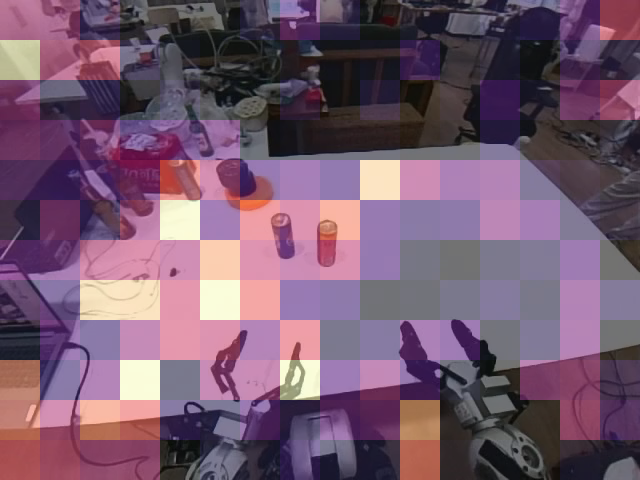}
        &
        \includegraphics[width=0.215\textwidth]{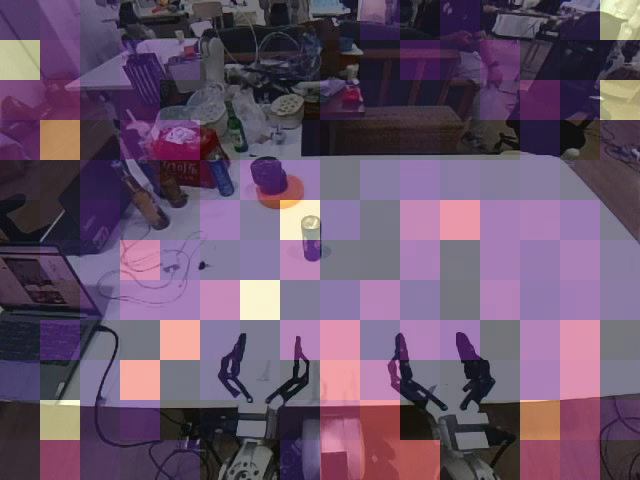}
        \\[3pt]

        \rotatebox{90}{\textbf{Robot B}}
        &
        \includegraphics[width=0.215\textwidth]{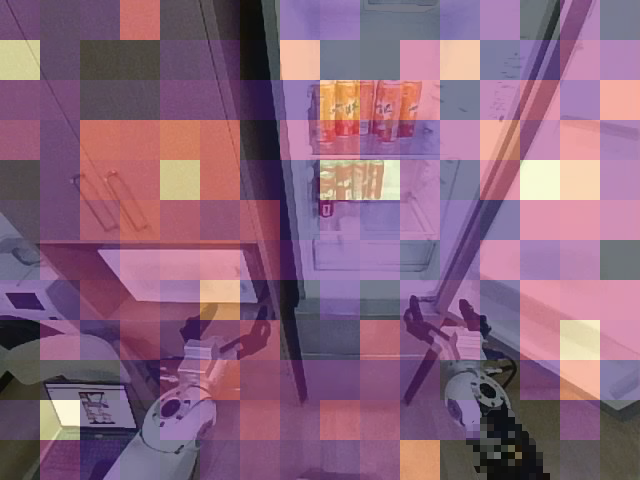}
        &
        \includegraphics[width=0.215\textwidth]{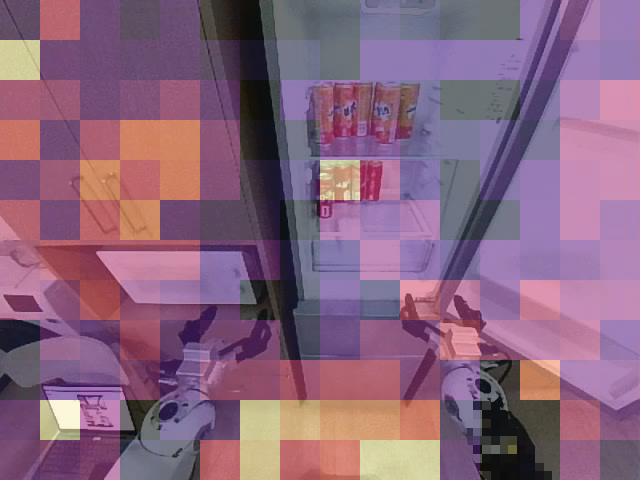}
        &
        \includegraphics[width=0.215\textwidth]{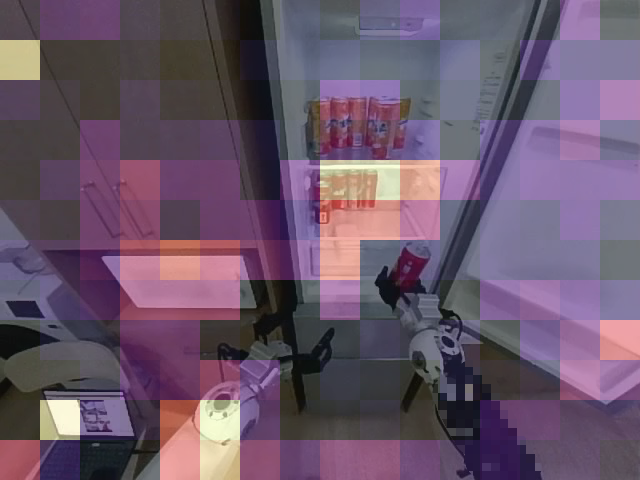}
        &
        \includegraphics[width=0.215\textwidth]{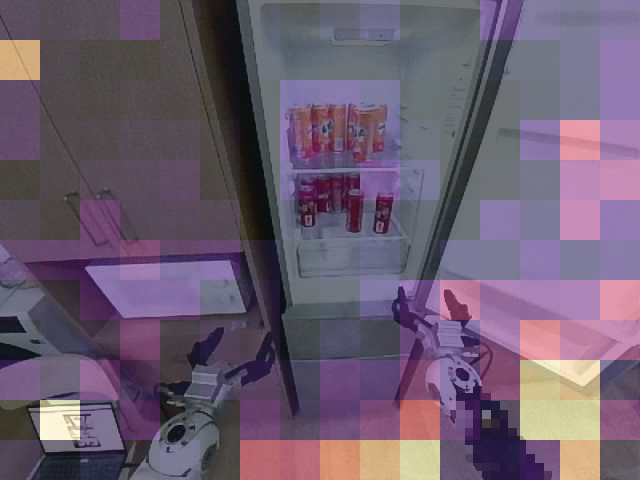}
    \end{tabular}

    \caption{
    Paired visual attention maps of Robots A and B during refrigerator restocking at four key moments. Each column shows the two robots' main-camera observations at the same timestep, with brighter regions indicating higher attention scores.
    }
    \label{fig:attention-score}
\end{figure*}

To further examine visual attention during collaboration, we visualize the patch-level attention scores extracted from the seventh layer of the VLA policy during action generation. Specifically, we pair the main-camera observations of Robots A and B at the same timestep and visualize their corresponding attention distributions. Regions with higher attention scores indicate the visual areas emphasized by each policy during action generation. By comparing the two views at the same collaborative moment, we examine whether the robots attend to complementary task-relevant information, such as target objects and interaction regions.

As illustrated in Fig.~\ref{fig:attention-score}, Robot A first opens the refrigerator and shares contextual information about the target placement locations with Robot B through the shared context tokens, with Fanta assigned to the upper shelf and Coke to the lower shelf. At $t_1$, the refrigerator and the Fanta bottle on the table receive higher attention, corresponding to the Fanta restocking stage. At $t_2$ and $t_3$, the highlighted regions shift toward the refrigerator and the Coke bottle on the table as the task progresses to Coke restocking. By $t_4$, neither the Fanta nor the Coke inside the refrigerator exhibits prominent attention responses, and no salient activation remains on the table, corresponding to the stage of checking for remaining relevant drinks. These observations suggest that the two robots exhibit complementary visual attention patterns from their respective viewpoints, with attention distributions changing consistently with task progression.

Overall, the task demonstrations in Fig.~\ref{fig:more_task_illustration} illustrate the applicability of shared context across different dual-robot collaborative scenarios, while the attention visualizations in Fig.~\ref{fig:attention-score} provide additional insights into how the robots attend to task-relevant visual information across different stages of collaboration. These attention maps serve as qualitative diagnostic evidence rather than direct causal explanations of how specific visual regions or shared context tokens influence action generation. Together with the quantitative comparisons and ablation studies, these qualitative observations help characterize the role of shared context in supporting multi-robot collaborative decision-making.

\section{Related Work}
\label{sec:related}

\subsection{Vision-Language-Action Models}
Recent Vision-Language-Action (VLA) models have advanced generalist manipulation through pretrained semantic representations, diverse training data, and multi-modal learning objectives. Pretrained vision-language backbones enable semantic knowledge transfer to robot control~\citep{rt_2,driess2023palm,pi_0,pi_0_5}, while large-scale, cross-embodiment training extends task and platform coverage~\citep{o2024open,team2024octo,kim2024openvla,liu2024rdt,li2024cogact}. Multi-modal co-training and auxiliary visual guidance further strengthen perception--action grounding~\citep{brohan2022rt,pi_0_5,li2025controlvla}. Recent approaches also introduce intermediate representations and explicit reasoning to bridge semantic understanding and action generation~\citep{zawalski2024robotic,zhao2025cot,lee2025molmoact,sun2024emma,huang2025thinkact,lumo1,lumo2}.
Despite training across diverse embodiments and tasks, these policies typically control a single robot using its local visual observations, language instructions, and proprioceptive state. Helix~\citep{helix} extends this paradigm by using shared model weights to control two humanoid robots in a collaborative, long-horizon manipulation task. While this demonstrates the potential of generalist VLAs for multi-robot collaboration, how to organize, compress, and exchange task-relevant perceptual information remains insufficiently explored as a design problem.
Our work addresses this cross-robot perceptual interface by encoding each robot's multi-camera observations into compact, task-conditioned latent representations that directly condition its partner's action policy. This formulation makes perceptual context sharing an explicit component of collaborative policy learning.

\subsection{Task Orchestration for Multi-Robot Collaboration}
LLM- and VLM-based agents are able to support multi-robot collaboration through task decomposition, capability matching, skill invocation, execution monitoring, and failure recovery. By incorporating execution feedback, these systems adapt their plans and coordination strategies to environmental changes and unexpected outcomes. InstAgent~\citep{instagent} extends this paradigm with recursive self-improvement (RSI) for heterogeneous robot teams. Its agentic behavior trees encode capability allocation, inter-robot dependencies, and task handover conditions. Following an execution failure, the system revises the relevant collaboration subtree, validates the revision through re-execution, and retains successful modifications for subsequent reuse.
These approaches primarily address task orchestration and execution management through high-level coordination interfaces, using shared memory and iterative refinement to improve task allocation, execution ordering, and failure recovery. Our work investigates the complementary problem of cross-robot perceptual context sharing during policy execution. Agentic orchestration and RSI can thus manage high-level coordination, while our framework supplies shared perceptual context for fine-grained collaborative action generation.

\subsection{Collaborative Perception}
Collaborative perception mitigates partial observability by enabling agents to exchange complementary information. In autonomous driving, V2VNet~\citep{v2vnet} shares compressed intermediate features for perception and motion prediction, while When2com~\citep{when2com} and Where2comm~\citep{where2comm} improve communication efficiency through learned agent selection and spatially selective feature exchange, respectively. These approaches primarily optimize communication for perceptual objectives rather than downstream action generation. In embodied manipulation, Sequential Asymmetric Imitation~\citep{sai} learns coordinated behaviors without explicit inter-robot communication, whereas MAAP~\citep{maap} integrates complementary wrist-camera observations for collaborative control.
Recent approaches explore explicit cross-robot information exchange. LatentToM~\citep{latenttom} learns individual and team-consensus representations for explicit communication and implicit partner-state inference. DuoMind~\citep{duomind} coordinates VLA policies through semantic information from a VLM orchestrator, while Latent Telepathy~\citep{latenttelepathy} shares frozen perceptual latents to provide task-relevant information unavailable from local observations. Our work learns compact, task-conditioned visual representations that directly condition a partner's VLA policy. Action supervision from the receiving robot jointly optimizes the context encoder and policy, aligning shared perceptual representations with collaborative action generation under a limited representation budget.

\subsection{Predictive and Progress-Aware Representation Learning}
Predictive representation learning models environmental dynamics through future-state prediction in latent space. V-JEPA2~\citep{vjepa2} learns dynamics through joint-embedding prediction and supports planning with an action-conditioned latent world model. FLARE~\citep{flare} aligns policy features with future observation representations, while LaWAM~\citep{lawam} uses compact latent visual subgoals to guide VLA policies. These approaches introduce predictive information into robot control without requiring pixel-level future generation.
Task progress provides a complementary signal for structuring embodied representations. XIRL~\citep{xirl} learns progress-sensitive embeddings from cross-embodiment videos and derives rewards from latent distances to goal states. VIP~\citep{vip} and LIV~\citep{liv} exploit temporal structure and value-based objectives for representation learning and reward modeling, with LIV additionally incorporating language goals. PROGRESSOR~\citep{progressor} estimates progress relative to initial and goal observations, while ProgVLA~\citep{progvla} and ProgressVLA~\citep{progressvla} incorporate progress prediction or guidance into long-horizon VLA policies.
Whereas these methods primarily use progress for reward modeling or individual-robot control, our work uses it to structure shared perceptual representations for multi-robot collaboration. A partner's execution phase can inform whether a robot should wait, proceed, or coordinate a handover. We combine task-oriented future prediction with soft task-progress correspondence learning to preserve predictive visual information while promoting inter-task discriminability and intra-task temporal continuity. Progress serves only as a training signal, rather than an explicit prediction or communication target. The resulting compact representations directly condition a partner's VLA policy, linking predictive and progress-aware representation learning to collaborative action generation.
\section{Conclusion, Limitations and Future Work}
\label{sec:conclusion}

We propose a two-stage active context sharing framework for multi-robot collaboration, in which each robot encodes its multi-camera observations and task instruction into a compact latent representation that is directly integrated into its partner’s VLA policy. The first stage learns shared representations that capture task-relevant information and execution-state structure, while the second stage further aligns these representations with collaborative decision-making through downstream action supervision. Extensive real-world dual-robot experiments demonstrate that the proposed framework provides decision-relevant information unavailable from local observations, improves target selection under visual occlusion, and enables bidirectional sharing of target identity and execution progress in alternating leader–follower tasks. Further ablations and communication-budget analyses validate the effectiveness of the two-stage training strategy and joint optimization, demonstrating that compact task-oriented tokens can achieve performance comparable to high-bandwidth generic visual features at substantially lower communication cost.

Although our experiments demonstrate the effectiveness of the proposed framework, the semantic content of the shared tokens remains underexplored. In particular, how visual semantics, task states, and execution-stage information are encoded in the latent representations, and how these representations influence the receiver's subsequent decisions and actions, remain open questions for further investigation.

Future work will explore extensions to larger robot teams, asynchronous and selective information sharing, and multi-modal collaborative representations that incorporate temporal memory, tactile sensing, and proprioception. More broadly, our results point toward a paradigm in which multi-robot collaboration moves beyond exchanging high-level commands or generic visual features, enabling robots to actively distill and communicate compact, task-relevant context shaped jointly by the sender's task objectives and the receiver's decision requirements.

\section{Contributions}
\label{sec:contribution}
\begin{itemize}
    \item  \textbf{Contributors:} Baifu Huang, Shilin Fang
    \item  \textbf{Project Lead:} Jianan Wang
\end{itemize}

\bibliography{main}
\clearpage
\end{document}